\documentclass[sigconf,natbib=true]{acmart}

\usepackage{booktabs}
\usepackage{amsfonts}
\usepackage{algorithm}
\usepackage{algorithmic}
\usepackage{threeparttable}  
\usepackage{multirow}
\usepackage{subfigure}
\usepackage{amsmath,bm}
\usepackage[symbol]{footmisc}

\copyrightyear{2022}
\acmYear{2022}
\setcopyright{acmcopyright}
\acmConference[CIKM '22] {Proceedings of the 31st ACM International Conference on Information and Knowledge Management}{October 17--21, 2022}{Atlanta, GA, USA.}
\acmBooktitle{Proceedings of the 31st ACM International Conference on Information and Knowledge Management (CIKM '22), October 17--21, 2022, Atlanta, GA, USA}
\acmPrice{15.00}
\acmISBN{978-1-4503-9236-5/22/10}
\acmDOI{10.1145/3511808.3557602}

\begin{document}

\title{GDA-HIN: A Generalized Domain Adaptive Model across Heterogeneous Information Networks}


\author{Tiancheng Huang}
\authornote{Both authors contributed equally to this research.}

\affiliation{%
\institution{Zhejiang University}
\country{}}
\affiliation{%
  \institution{Westlake University}
  \country{}
}
\affiliation{%
  \institution{Westlake Institute for Advanced Study, Hangzhou, China}
  \country{}
}

\email{huangtiancheng@westlake.edu.cn}

\author{Ke Xu}
\authornotemark[1]
\affiliation{%
  \institution{Westlake University}
  \country{}
}
\affiliation{%
  \institution{Westlake Institute for Advanced Study, Hangzhou, China}
  \country{}
}
\email{xuke@westlake.edu.cn}

\author{Donglin Wang}
\authornote{Corresponding author.}
\affiliation{%
  \institution{Westlake University}
  \country{}
}
\affiliation{%
  \institution{Westlake Institute for Advanced Study, Hangzhou, China}
  \country{}
}
\email{wangdonglin@westlake.edu.cn}


\begin{abstract}
Domain adaptation using graph-structured networks learns label-discriminative and network-invariant node embeddings by sharing graph parameters. Most existing works focus on domain adaptation of homogeneous networks. The few works that study heterogeneous cases only consider shared node types but ignore private node types in individual networks. However, for given source and target heterogeneous networks, they generally contain shared and private node types, where private types bring an extra challenge for graph domain adaptation. In this paper, we investigate Heterogeneous Information Networks (HINs) with both shared and private node types and propose a Generalized Domain Adaptive model across HINs (GDA-HIN) to handle the domain shift between them. GDA-HIN can not only align the distribution of identical-type nodes and edges in two HINs but also make full use of different-type nodes and edges to improve the performance of knowledge transfer. Extensive experiments on several datasets demonstrate that GDA-HIN can outperform state-of-the-art methods in various domain adaptation tasks across heterogeneous networks.
\end{abstract}



\maketitle

\section{Introduction}
\label{sec:introdution}
\begin{figure}[t]
	\centering
	\includegraphics[scale = 0.27]{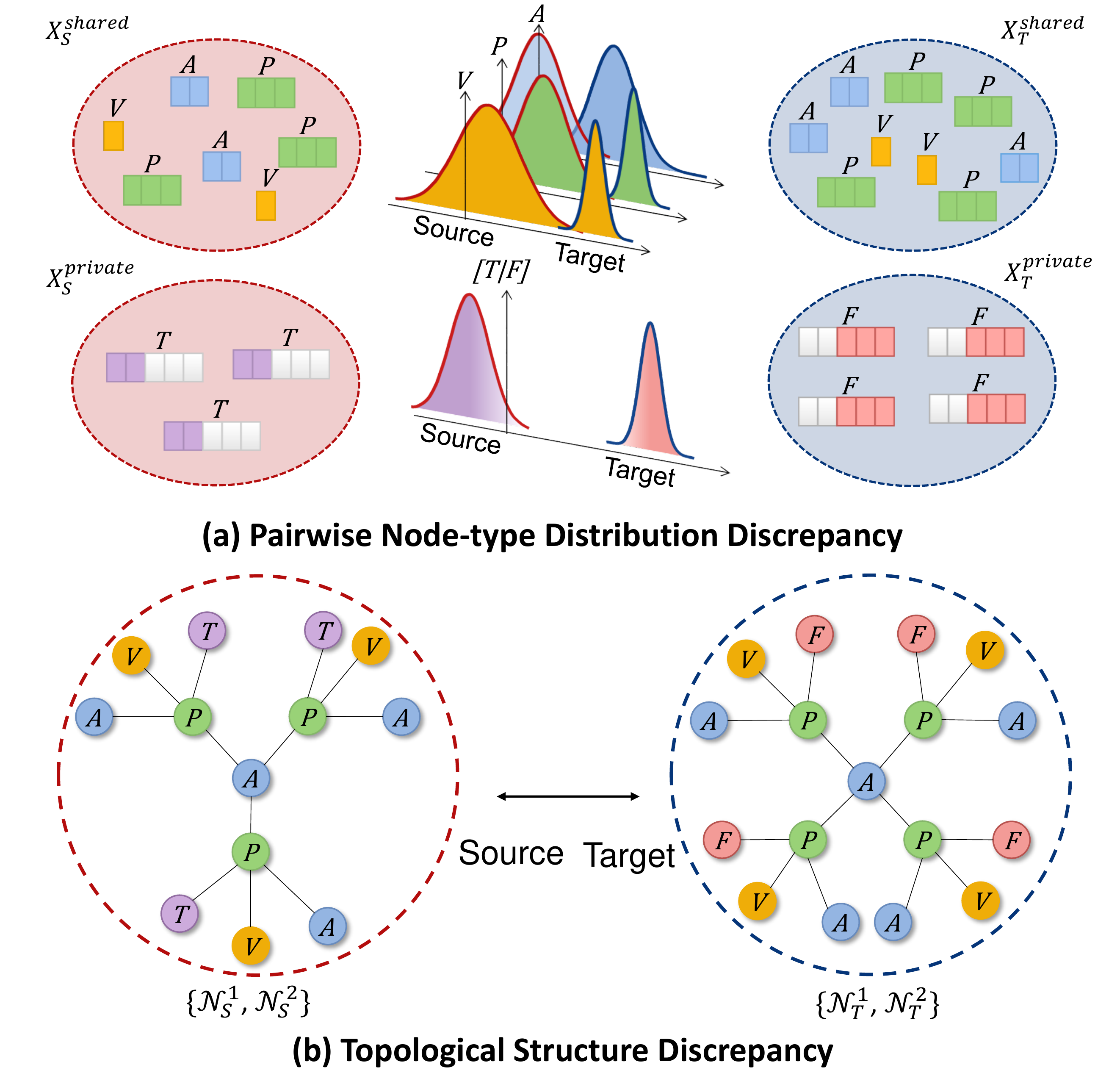}
	\vspace{-10pt}
	\caption{Challenges on DA across HINs: (a) feature and (b) structure discrepancies between source and target HINs.}
	\vspace{-10pt}
	\label{fig:challenges}
\end{figure}

Domain Adaptation (DA) aims to learn transferable representations for problems those sample and label spaces remain unchanged, but the probability distribution is different \cite{kouw2018an,zhuang2019a,long2015learning,fu2017domain}.
Recently, DA across graphs has been beginning to draw much attention \cite{shen2019network,shen2020adversarial,zhang2019dane:,wu2020unsupervised}.
It aims to transfer the knowledge learned from a source network to a target network by learning label-discriminative and network-invariant node embeddings.
However, existing methods mainly focus on the problem of homogeneous networks.
In this paper, we study a generalized domain adaptation across Heterogeneous Information Networks (HINs), where nodes are fully labeled in the source network while completely unlabeled in the target network.
We intend to minimize the domain shift and transfer the model trained on the source network to the target network.

However, this objective faces several challenges: 
1) First, each HIN is composed of multiple different types of nodes and edges. 
For example, the source network DBLP \cite{ji2010graph} in Fig. \ref{fig:challenges}(a) contains four types of nodes: papers (P), authors (A), venues (V), and terms (T), as well as different types of edges between them.
This means that different types of nodes and edges are in different semantic spaces and have different data distributions, which results in three discrepancies between \emph{pairwise node-type} distributions (See Fig. \ref{fig:challenges} (a.top)). 
2) Second, the node and edge types in two heterogeneous networks are not identical in general.
For instance, DBLP and MAG \cite{sinha2015overview} citation networks have some private-type nodes and edges in their networks (i.e., term nodes for DBLP, and field (F) nodes for MAG).
Because private node types exist across the networks (See Fig. \ref{fig:challenges} (a.bottom)), a common semantic space needs to be built before they are aligned to capitalize on the information about such type nodes.
For topological structures from two different domains, there is a discrepancy to be aligned for the center node embedding even after all distributions of node types have been aligned (See Fig. \ref{fig:challenges} (b)).
However, previous work \cite{yang2020domain} only aligns the shared type node features but ignores that each network has its characteristics.

\begin{figure*}[t]
	\centering
	\vspace{-5pt}
	\includegraphics[scale = 0.42]{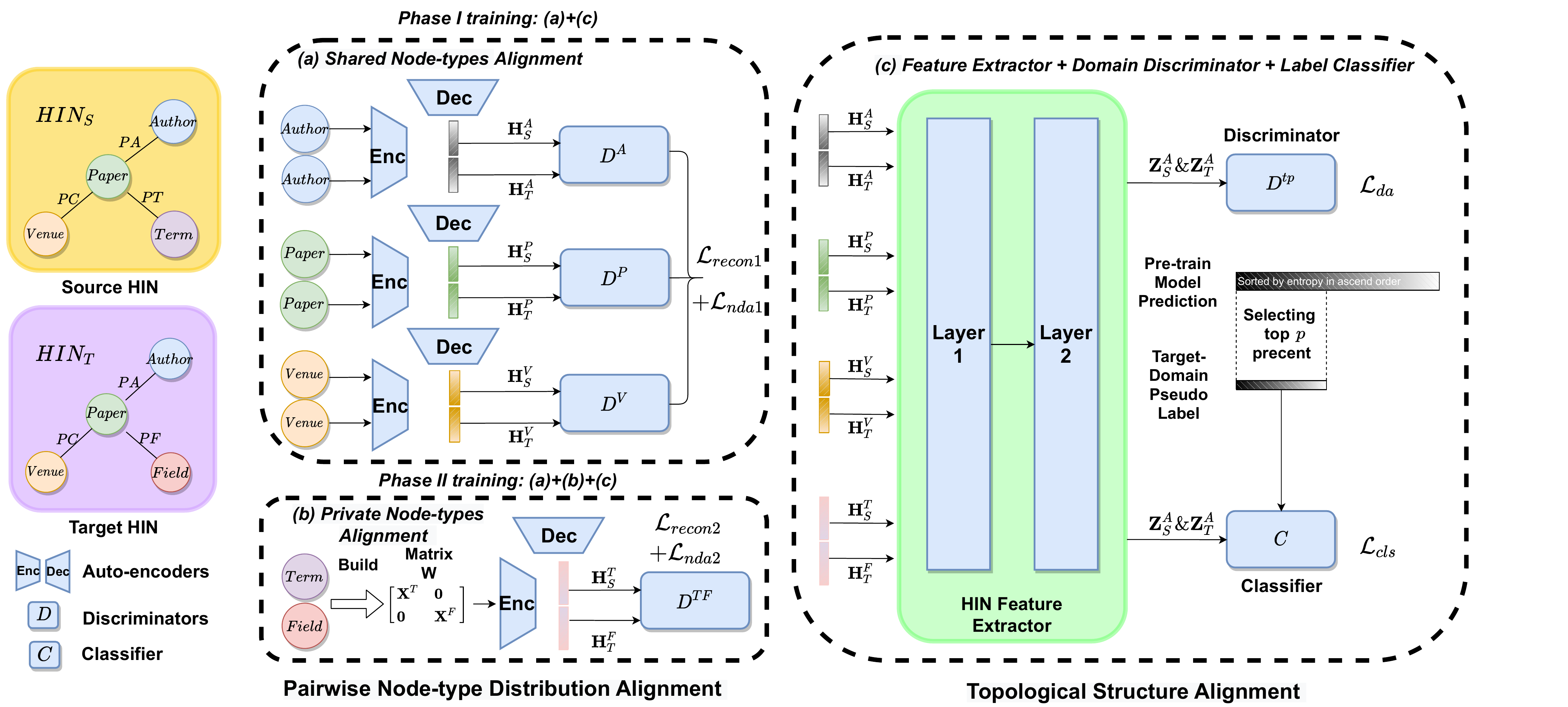}
	\caption{The framework of GDA-HIN. See more details in Methodology (\S \ref{sec:methology}).}
	\label{fig:framework}
\end{figure*}

To address these problems, we propose a  Generalized Domain Adaptive model across HINs (GDA-HIN).
Several independent discriminators are used to align the embeddings in different semantic spaces. 
Specifically, our model firstly constructs auto-encoders and domain discriminators to align the node features.
Secondly, we align the topological structure in graph embedding space using an automatically learned meta-path feature extractor (e.g., HGT \cite{hu2020heterogeneous}) and a domain discriminator.
A low-rank matrix completion method is adopted to handle private node types.
Our model combines the low-rank matrix completion method with the heterogeneity of HIN.
In this way, GDA-HIN can get rid of the precondition of matrix completion method, that corresponding instances \cite{xiao2013novel}, or a few labeled target domain data \cite{li2019heterogeneous}.
The contributions of this work:

1) To the best of our knowledge, we are the first to investigate the generalized domain adaptation across HINs that contain both share and private node types.
Thus, it is a more general situation on source and target HINs for real-world scenarios.

2)  We systematically analyze the challenges of DA across HINs. 
Based on the above analyses, we propose a novel GDA-HIN by designing pairwise node-type distribution alignment and topological structure alignment to accomplish DA across HINs.

3) We conduct extensive experiments on the three citation networks and six groups of cross-network tasks for node classification to evaluate the performance of domain adaptation. 
The experiments show that GDA-HIN outperforms state-of-the-art baselines.


\section{Methodology}
\label{sec:methology}

\subsection{Problem Definition}
Given a fully-labeled source $HIN_S=\left(\mathcal{V}_S^{l},\mathcal{E}_S, \mathcal{A}_S, \mathcal{R}_S\right)$ and a fully-unlabeled target $HIN_T=\left(\mathcal{V}_T^{u},\mathcal{E}_T, \mathcal{A}_T, \mathcal{R}_T\right)$, where $\mathcal{V}$ and $\mathcal{E}$ are node and edge set, $\mathcal{A}$ and $\mathcal{R}$ denote the sets of node and edge types. 
$\mathcal{A}_S$ and $\mathcal{A}_T$ not only share same node types but also have their own private node types, also leading to the difference between $\mathcal{R}_S$ and $\mathcal{R}_T$. $\mathcal{V}_S^{l}$ represents a set of labeled nodes while $\mathcal{V}_T^{u}$ represents a set of unlabeled nodes.
The transferable classification aims to predict the label on $HIN_T$ with label information from $HIN_S$.

\subsection{Pairwise Node-type Distribution Alignment}
To align each semantic component independently between source and target domains, we construct  \emph{pairwise node-type} auto-encoders and domain discriminators before aggregation and updating:

\noindent
\textbf{1) Shared Node-type Distribution Alignment:}
Assuming there are $K_1$ shared node-type pairs between source and target domains, the \emph{pairwise node-type} auto-encoder $AE^{k_1},{k_1}=1,...,K_1$ is used to encode and decode the node features of ${k_1}$-th-type nodes.
A reconstruction loss is used to constrain $AE^{k_1}$'s projection retaining semantic information:
\begin{equation} \small
\label{equa:loss_MSE}
\mathcal{L}_{\text {recon1}} = \sum MSE\left(\mathbf{X}^{k_1},\hat{\mathbf{X}}^{k_1}\right),
\end{equation}
\noindent where $\mathbf{X}^{k_1}$ is the node feature of the ${k_1}$-th shared node type for both domains, $\hat{\mathbf{X}}^{k_1}$ is the corresponding reconstructed feature, and $MSE$ represents the mean square error.
Then, we construct a \emph{pairwise node-type} domain discriminator following with a Gradient Reversal Layer (GRL) \cite{ganin2016domain} for the features of individual-type nodes separately.
By minimizing the domain adversarial similarity loss, 
the encoder is trained to make similar its output processed from both domains while the domain discriminator learns to identify the domain of the encoder's output.
The domain discriminator loss for all discriminators $D^{k1}$ and auto-encoders $AE^{k1}$, is formulated as:
\begin{align}
\label{equa:loss_nda1}
\mathcal{L}_{nda1} = \sum \mathbb{E}_{x \in \textbf{X}_S^{k_1}} \left[\textbf{{\rm log}} \left(1-D^{k_1}\left(AE^{k_1}\left(x\right)\right)\right)\right] \nonumber \\
+\mathbb{E}_{x \in \textbf{X}_T^{k_1}}\left[\textbf{{\rm log}} D^{k_1}\left(AE^{k_1}\left(x\right)\right)\right],
\end{align}
\noindent
where $D^{k_1}$ is the discriminator for the ${k_1}$-th-type nodes.

\noindent
\textbf{2) Private Node-type Distribution Alignment:} 
Unlike shared-node types, private-node types contain many unknown values.
Reconstruction constraints applied to private-node types recover not only the observed values but also unknown parts.
Meanwhile, the private-node types like term and field are semantically relevant \cite{mikolov2013efficient}, which makes the unobserved type's embedding contain plenty of linear dependent columns.
Hence, the problem of recovery unobserved type's embedding turns into a low-rank matrix completion problem, which can be formulated to minimize the nuclear norm under the constraint of reconstruction loss \cite{candes2009exact}.
For private node-type pairs, we recover the missing value and get a matrix $\hat{\mathbf{W}}$:
\begin{equation}
\label{equa:matrix_recover}
\mathbf{W}=\left[\begin{array}{ll}
\textbf{X}_{S} & \textbf{0} \\
\textbf{0} & \textbf{X}_{T}
\end{array}\right] \stackrel{recover}{\Longrightarrow}
\hat{\mathbf{W}}=\left[\begin{array}{ll}
\hat{\textbf{X}}_{S} & \hat{\textbf{U}}_{S} \\
\hat{\textbf{U}}_{T} & \hat{\textbf{X}}_{T}
\end{array}\right].
\end{equation}

In $\hat{\mathbf{W}}$ above, $\hat{\mathbf{X}}_S$ and $\hat{\mathbf{X}}_T$ represent the recovered observed elements, while $\hat{\mathbf{U}}_S$ and $\hat{\mathbf{U}}_T$ represent the recovered unobserved elements.
To recover $\hat{\mathbf{W}}$, we minimize the loss $\mathcal{L}_{\text {recon2}}$:
\begin{align}
\label{equa:loss_recon2}
\mathcal{L}_{\text {recon2}} &=\sum MSE\left(\mathbf{X}^{k_2},\hat{\mathbf{X}}^{k_2}\right) +\delta\mathcal{R}\left(\hat{\mathbf{W}}\right),
\end{align}
\noindent 
where $\mathbf{X}^{k_2}$ is the node feature of the $k_2$-th private node type, $\hat{\mathbf{X}}^{k_2}$ is corresponding observed part of recovered embedding, $K_2$ is the number of the private-type pairs.
$\mathcal{R}(\hat{\mathbf{W}})$ is a regularization term, where $\mathcal{R}(*)$ denotes nuclear norm operation, and $\delta >0$ is a trade-off parameter. 
Under the reconstruction constraint, the encoder's output can retain enough semantic information for the private-type pairs.
The loss of discriminators for $k_1$ shared- and $k_2$ private-type pairs is $\mathcal{L}_{nda} = \mathcal{L}_{nda1} + \mathcal{L}_{nda2}$, where $\mathcal{L}_{nda2}$ is similar as Eq. (\ref{equa:loss_nda1}).

\subsection{Topological Structure Alignment} 
In this subsection, we further elaborate on how to align the topological structure of source and target networks.

\noindent
\textbf{Representation of $\mathbf{h}$-hop:} 
We choose the advanced HIN model HGT \cite{hu2020heterogeneous} as feature extractor $G$, which can learn embeddings of $\mathbf{h}$-hop structure ($\{\mathcal{N}^1_{(v)},...,\mathcal{N}^h_{(v)}\}$), that is $\mathbf{h}$-hop embedding, to capture network topology information. 
In GDA-HIN, nodes from the source and target networks are encoded via a feature extractor with shared learnable parameters.
However, there are still $\mathbf{h}$-hop structure discrepancies between both domains.
Then, we adopt the domain alignment on the output of HGT, which aligns the data distribution in the embedding space of $\mathbf{h}$-hop structure.

\noindent
\textbf{Topological Domain Discriminator:} 
After extracting each nodes' h-hop structure embedding by feature extractor $G$, which is a 2-layer HGT, another domain adversarial discriminator $D^{tp}$ is implemented to minimize the topological structure discrepancy, and its loss $\mathcal{L}_{da}$ defined as:
\begin{align} 
\label{equa:loss_da}
\mathcal{L}_{da} = \mathbb{E}_{x \in \boldsymbol{H_S}}\left[\textbf{{\rm log}} \left(1-D^{tp}\left(G\left(x\right)\right)\right)\right] \\ \nonumber
+\mathbb{E}_{x \in \boldsymbol{H_T}}\left[\textbf{{\rm log}} D^{tp}\left(G\left(x\right)\right)\right],
\end{align}%
\noindent
where $\mathbf{H}_S$ and $\mathbf{H}_T$ represent the outputs of encoders in source and target domains, respectively.


\noindent
\textbf{Domain-invariant Classifier:} 
The classifier $C$ is used to to predict the label,
and its loss $\mathcal{L}_{cls}$ is defined as:
\begin{align}
\label{equa:loss_t_cls}
\mathcal{L}_{cls} =-\frac{1}{N^l}\mathop{\Sigma}_ {i=0}^{N^l}{Y^i log\left(\hat{Y}^i\right)} + \zeta  tr\left(\mathbf{H}^\top \mathbf{L^gH}\right),
\end{align}
\noindent
where $\zeta$ is a balance parameter, $N^l$ represents the number of labeled source- and target-domain nodes, and $\hat{Y}^i$ denotes the $i$-th node's prediction.
The regularization term in Eq.(\ref{equa:loss_t_cls}) is defined as $tr(\mathbf{H}^\top \mathbf{L^gH})$, where $\mathbf{H}$ represents the hidden state of auto-encoders for all private-type nodes and $\mathbf{L^g}$ denotes the graph Laplacian matrix for all private-type nodes.
In particular, the graph Laplacian matrix is formulated as $\mathbf{L^g} = \begin{pmatrix} \mathbf{L_S} & \mathbf{0} \\ \mathbf{0} & \mathbf{L_T} \end{pmatrix}$, where $\mathbf{L_S}$ and $\mathbf{L_T}$ are the Laplacian matrices of source and target domains computed according to the adjacency matrices. 
In the matrix completion module, there are unobserved elements involved in the computation of private-type nodes' embeddings, which are not constrained by reconstruction loss.
We utilize graph Laplacian matrix to smooth their embedding over the graph, relying on the assumption that connected nodes in the graph are likely to share the same label \cite{kipf2016semi}.

\subsection{Optimization}
\noindent 
\textbf{Phase \uppercase\expandafter{\romannumeral1} Training:}
To minimize the nuclear norm under the constraint of  reconstruction loss Eq.(\ref{equa:loss_recon2}), GDA-HIN trains a \emph{Phase \uppercase\expandafter{\romannumeral1} Training model} on the shared node types of two networks to yeild pseudo labels, and the overall objective function is composed of the following four components:
\begin{align}
\label{equa:l_stage1}
\mathcal{L}_{p1} =  \mathcal{L}_{cls}+\alpha \mathcal{L}_{recon1} + \beta\mathcal{L}_{nda1} + \gamma\mathcal{L}_{da},
\end{align}%

\noindent where $\alpha$, $\beta$ and $\gamma$ are hyper-parameters. 

When there are only shared node types, Eq.(\ref{equa:loss_t_cls}) degenerates to cross-entropy loss,
$
\mathcal{L}_{cls} =-\frac{1}{N_S}\mathop{\Sigma}_ {i=0}^{N_S}{Y_S^ilog\left(\hat{Y}_S^i\right)},
$
where $N_S$ denotes the number of source-domain nodes.

\noindent 
\textbf{Phase \uppercase\expandafter{\romannumeral2} Training:}
For phase \uppercase\expandafter{\romannumeral2} training, we select some predictions by \emph{Phase \uppercase\expandafter{\romannumeral1} Training model} as pseudo labels for nodes in target domain. 
Under the previous phase's guidance, GDA-HIN considers both shared- and private- node types from two networks, and the overall optimization objective is:
\begin{align}
\label{equa:l_stage2}
\mathcal{L}_{p2} = \mathcal{L}_{cls}+\alpha \left(\mathcal{L}_{recon1}+\mathcal{L}_{recon2}\right)
+ \beta \mathcal{L}_{nda} + \gamma\mathcal{L}_{da}.
\end{align}

\section{Experiments}
\label{sec:experiments}

\begin{table}[H] \small
	\centering
	\caption{Dataset statistics.}
	\label{tab:dataset_statistics}
	\begin{tabular}{lcccccccccc}
		\hline
		\multirow{2}{*}{\textbf{Dataset}} & \multicolumn{3}{c}{\textbf{Shared}} & \multicolumn{1}{c}{\textbf{Private}} & \multicolumn{2}{c}{\textbf{Shared}} & \multicolumn{1}{c}{\textbf{Private}} \\ \cline{2-4} \cline{6-7}
		& \textbf{P}   & \textbf{A}   & \textbf{V}   & \textbf{T/K/F}    & \textbf{P-A}             & \textbf{P-V}             & \textbf{P-T/K/F}            \\ \hline
		DBLP & 14,328      & 4,057        & 20         & 2,517         & 19,645             & 14,328             & 8,647      \\ 
		Aminer                   & 7,212       & 4,696        & 16         & 7,323     & 13,796             & 7,212    & 22,568   \\ 
		MAG                      & 6,206       & 5,861        & 20         & 2,370       & 12,615             & 6,206              & 8,701      \\ 
		\hline
	\end{tabular}
\end{table}

\subsection{Experiment Settings}
\label{appendix:datasets}
\textbf{Datasets:}
\textbf{DBLP} \cite{ji2010graph}. We extract a subset of DBLP which contains 14,328 papers (P), 4,057 authors (A), 20 venues (V), 2,517 terms (T), and the edges between nodes.   
\textbf{Aminer} \cite{tang2008arnetminer}. We extract a subset of Aminer by selecting papers published from the year 2004 to 2008, which contains 7,212 papers, 4,696 authors, 16 venues, 7,323 keywords (K), and the edges between nodes. 
\textbf{MAG} \cite{sinha2015overview}. Here we extract a subset of MAG with publish date between the year 2017 and 2019, which contains 6,206 papers, 5,861 authors, 20 venues, 2,370 fields (F), and the edges between nodes. 
We categorize authors according to their research areas: \emph{Database}, \emph{Data Mining}, \emph{Artificial Intelligence}, and \emph{Information Retrieval}.
Summary statistics of the datasets are displayed in Table \ref{tab:dataset_statistics}. 


\noindent
\textbf{Baselines.}
We compare with the following baselines:
\textbf{GCN+GRL}. The model adopts homogeneous graph-based methods GCN \cite{kipf2016semi} as feature extractors and take GRL \cite{ganin2016domain} as domain adaptation framework. 
\textbf{UDA-GCN}. The method \cite{wu2020unsupervised} adopts a multi-channel GCN with a weight sharing strategy and takes GRL as the domain adaptation framework. 
It effectively maintains local consistency and global consistency of the graphs. 
\textbf{HAN+GRL}. This model adopts HIN-based method HAN \cite{wang2019heterogeneous} as feature extractor, and takes GRL as the domain adaptation framework.

\begin{table}[t] \small
	\centering
	\caption{Node classification accuracy (\%) comparisons on six cross tasks. (D: DBLP; A: Aminer; M: MAG)}
	\begin{tabular}{lccccccc}
		\hline
		\textbf{Methods} & \textbf{D$\rightarrow$A} & \textbf{D$\rightarrow$M} & \textbf{A$\rightarrow$D} & \textbf{A$\rightarrow$M} & \textbf{M$\rightarrow$D} & \textbf{M$\rightarrow$A} \\ \hline 
		GCN+GRL &  51.30  &  35.56  &  42.17  &  33.39   &  42.15  &  51.21             \\ 
		UDA-GCN &  55.86  &  36.14  &  47.10  &  34.55   &  46.86  &  51.24  	\\  
		HAN+GRL & 56.73 & 34.98 & 45.77 & 33.66 & 45.52 & 51.34          \\ \hline
		HGT ($w/o$ DA)  & 34.11 & 28.83 & 30.74 & 29.14 & 27.24 & 31.35  \\ 
		GDA-HIN$^{w/o \ P}$ & 53.13 &	36.72 &	41.41 &	36.33 & 47.83 & 51.36  \\ 
		GDA-HIN$^{w/o \ T}$ & 54.10 &	32.42 &	44.73 & 35.00 &	48.02 &	46.29 \\ 
		GDA-HIN$^{w/ \ S}$ & 58.18    & 37.71    & 47.77    & 36.80     & 48.66     & 52.87              \\ 
		\hline
		GDA-HIN(ours) & \textbf{58.84}              & \textbf{39.45}              & \textbf{58.59 }             & \textbf{36.91}              & \textbf{50.39}              & \textbf{54.30}              \\ 
		\hline
	\end{tabular}
	\label{tab:experimental_result}
\end{table}

\subsection{Performance Comparison}
\label{subsec:performance_comparison}
Our model carries out six groups of cross-network tasks, and the node classification results are reported in Table \ref{tab:experimental_result}. We have the following primary observations:
(1) Compared with all baselines, the proposed GDA-HIN generally achieves the best performance. 
The results demonstrate the effectiveness of GDA-HIN.
(2) All HIN-based methods exhibit better performance than homogeneous methods (i.e., GDA-HIN vs. GCN+GRL). 
This result reveals that considering heterogeneity helps the DA. 
In essence, it validates that for the domain adaptation problem of HINs, it is essential to handle the heterogeneity-caused multiple semantic spaces.
(3) The GDA-HIN outperforms other heterogeneous graph-based methods, which demonstrates that our method can better address the discrepancies of both pairwise node-type distributions and topological structure.

\subsection{Ablation Study}
\label{subsec:ablation_study}
To gain insights about GDA-HIN, we study its variants:
GDA-HIN$^{w/o \ P}$, GDA-HIN$^{w/o \ T}$ and
GDA-HIN$^{w/ \ S}$.
Specifically, GDA-HIN$^{w/o \ P}$ \& GDA-HIN$^{w/o \ T}$ are two variants of GDA-HIN with the pairwise node-type distribution alignment and the topological structure alignment \textit{removed}, respectively.
GDA-HIN$^{w/ \ S}$ is only uses shared node types for \emph{phase \uppercase\expandafter{\romannumeral1} training}, and without using private node types,
to verify the effectiveness of private node types.

From Table \ref{tab:experimental_result}, we observe that GDA-HIN is better than GDA-HIN$^{w/o \ P}$, demonstrating the effectiveness of the pairwise node-type distribution alignment for domain shift.
Similarly, it is observed that GDA-HIN outperforms GDA-HIN$^{w/o \ T}$, verifying the effectiveness of the topological structure
alignment. 
Moreover, GDA-HIN outperforms GDA-HIN$^{w/ \ S}$ on all cross-domain tasks, indicating the effectiveness of considering private types in domain alignment. 
Note that GDA-HIN$^{w/ \ S}$ outperforms other baselines, it indicates our model only using share- node types gains competitive results.
In summary, GDA-HIN outperforms three variants, indicating that pairwise node-type alignment, which includes shared and private alignments, and topological structure alignment are indispensable.

\subsection{Visualization}
\label{sub:visualization}

\begin{figure}[t]
	\centering
	\subfigure[HGT ($w/o$ DA)]{\includegraphics[scale = 0.30]{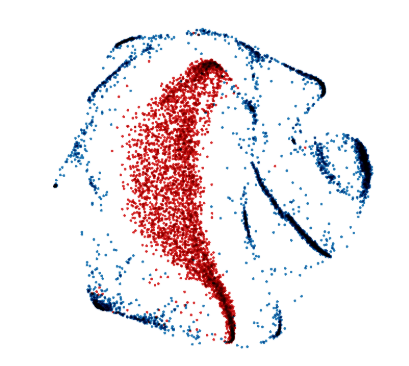}}
	\subfigure[GDA-HIN]{\includegraphics[scale = 0.30]{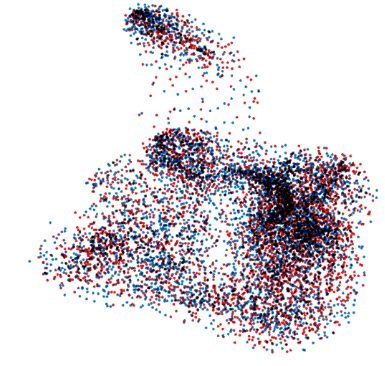}}
	\caption{The visualization: the data points of source and target domains are colored by red and blue color, respectively.}
	\label{fig:visualization}
\end{figure}

Due to the space limit, we report the t-SNE visualization of models' embeddings on Aminer$\rightarrow$DBLP as an illustration.
The results are shown in Fig. \ref{fig:visualization}, where the source and target domains are colored according to their own domains. 
From Fig. \ref{fig:visualization} (a) $\sim$ (b), we can find that the result of (a) HGT contains a tremendous distribution discrepancy due to $w/o$ DA. 
Apparently, the visualization of our (b) GDA-HIN performs best, where data points of source and target domains are evenly mixed and hardly separated, which means that the domain shift has been successfully minimized, and the node embeddings learned by GDA-HIN are actually domain-invariant.
The domain-invariant representations are beneficial to transferring from source domain to target domain.

\section{Conclusion}
\label{sec:conclusion}
In GDA-HIN, we achieve a better domain adaptation performance by jointly considering the pairwise node-type distribution alignment and topology structure alignment.
We adopt the matrix completion method to effectively deal with private-type nodes by projecting each private-type pair into a new common feature space.
The proposed scheme has been proven effective by experiments.

\section*{Acknowledgements}
This work was supported by NSFC General Program (Grant No. 62176215), and National Science and Technology Innovation 2030 - Major Project (Grant No. 2022ZD0208800). 

\clearpage
\bibliographystyle{ACM-Reference-Format}
\bibliography{sample-base}


\begin{thebibliography}{20}


\ifx \showCODEN    \undefined \def \showCODEN     #1{\unskip}     \fi
\ifx \showDOI      \undefined \def \showDOI       #1{#1}\fi
\ifx \showISBNx    \undefined \def \showISBNx     #1{\unskip}     \fi
\ifx \showISBNxiii \undefined \def \showISBNxiii  #1{\unskip}     \fi
\ifx \showISSN     \undefined \def \showISSN      #1{\unskip}     \fi
\ifx \showLCCN     \undefined \def \showLCCN      #1{\unskip}     \fi
\ifx \shownote     \undefined \def \shownote      #1{#1}          \fi
\ifx \showarticletitle \undefined \def \showarticletitle #1{#1}   \fi
\ifx \showURL      \undefined \def \showURL       {\relax}        \fi
\providecommand\bibfield[2]{#2}
\providecommand\bibinfo[2]{#2}
\providecommand\natexlab[1]{#1}
\providecommand\showeprint[2][]{arXiv:#2}

\bibitem[\protect\citeauthoryear{Cand{\`e}s and Recht}{Cand{\`e}s and
  Recht}{2009}]%
        {candes2009exact}
\bibfield{author}{\bibinfo{person}{Emmanuel~J Cand{\`e}s} {and}
  \bibinfo{person}{Benjamin Recht}.} \bibinfo{year}{2009}\natexlab{}.
\newblock \showarticletitle{Exact matrix completion via convex optimization}.
\newblock \bibinfo{journal}{\emph{Foundations of Computational mathematics}}
  \bibinfo{volume}{9}, \bibinfo{number}{6} (\bibinfo{year}{2009}),
  \bibinfo{pages}{717--772}.
\newblock


\bibitem[\protect\citeauthoryear{Fu, Nguyen, Min, and Grishman}{Fu
  et~al\mbox{.}}{2017}]%
        {fu2017domain}
\bibfield{author}{\bibinfo{person}{Lisheng Fu}, \bibinfo{person}{Thien~Huu
  Nguyen}, \bibinfo{person}{Bonan Min}, {and} \bibinfo{person}{Ralph
  Grishman}.} \bibinfo{year}{2017}\natexlab{}.
\newblock \showarticletitle{Domain adaptation for relation extraction with
  domain adversarial neural network}. In \bibinfo{booktitle}{\emph{Proceedings
  of the Eighth International Joint Conference on Natural Language Processing
  (Volume 2: Short Papers)}}. \bibinfo{pages}{425--429}.
\newblock


\bibitem[\protect\citeauthoryear{Ganin, Ustinova, Ajakan, Germain, Larochelle,
  Laviolette, Marchand, and Lempitsky}{Ganin et~al\mbox{.}}{2016}]%
        {ganin2016domain}
\bibfield{author}{\bibinfo{person}{Yaroslav Ganin}, \bibinfo{person}{Evgeniya
  Ustinova}, \bibinfo{person}{Hana Ajakan}, \bibinfo{person}{Pascal Germain},
  \bibinfo{person}{Hugo Larochelle}, \bibinfo{person}{Fran{\c{c}}ois
  Laviolette}, \bibinfo{person}{Mario Marchand}, {and} \bibinfo{person}{Victor
  Lempitsky}.} \bibinfo{year}{2016}\natexlab{}.
\newblock \showarticletitle{Domain-adversarial training of neural networks}.
\newblock \bibinfo{journal}{\emph{The journal of machine learning research}}
  \bibinfo{volume}{17}, \bibinfo{number}{1} (\bibinfo{year}{2016}),
  \bibinfo{pages}{2096--2030}.
\newblock


\bibitem[\protect\citeauthoryear{Hu, Dong, Wang, and Sun}{Hu
  et~al\mbox{.}}{2020}]%
        {hu2020heterogeneous}
\bibfield{author}{\bibinfo{person}{Ziniu Hu}, \bibinfo{person}{Yuxiao Dong},
  \bibinfo{person}{Kuansan Wang}, {and} \bibinfo{person}{Yizhou Sun}.}
  \bibinfo{year}{2020}\natexlab{}.
\newblock \showarticletitle{Heterogeneous graph transformer}. In
  \bibinfo{booktitle}{\emph{Proceedings of The Web Conference 2020}}.
  \bibinfo{pages}{2704--2710}.
\newblock


\bibitem[\protect\citeauthoryear{Ji, Sun, Danilevsky, Han, and Gao}{Ji
  et~al\mbox{.}}{2010}]%
        {ji2010graph}
\bibfield{author}{\bibinfo{person}{Ming Ji}, \bibinfo{person}{Yizhou Sun},
  \bibinfo{person}{Marina Danilevsky}, \bibinfo{person}{Jiawei Han}, {and}
  \bibinfo{person}{Jing Gao}.} \bibinfo{year}{2010}\natexlab{}.
\newblock \showarticletitle{Graph regularized transductive classification on
  heterogeneous information networks}. In \bibinfo{booktitle}{\emph{Joint
  European Conference on Machine Learning and Knowledge Discovery in
  Databases}}. Springer, \bibinfo{pages}{570--586}.
\newblock


\bibitem[\protect\citeauthoryear{Kipf and Welling}{Kipf and Welling}{2017}]%
        {kipf2016semi}
\bibfield{author}{\bibinfo{person}{Thomas~N Kipf} {and} \bibinfo{person}{Max
  Welling}.} \bibinfo{year}{2017}\natexlab{}.
\newblock \showarticletitle{Semi-supervised classification with graph
  convolutional networks}.
\newblock \bibinfo{journal}{\emph{The International Conference on Learning
  Representations}} (\bibinfo{year}{2017}).
\newblock


\bibitem[\protect\citeauthoryear{Kouw and Loog}{Kouw and Loog}{2018}]%
        {kouw2018an}
\bibfield{author}{\bibinfo{person}{Wouter~M Kouw} {and} \bibinfo{person}{Marco
  Loog}.} \bibinfo{year}{2018}\natexlab{}.
\newblock \showarticletitle{An introduction to domain adaptation and transfer
  learning}.
\newblock \bibinfo{journal}{\emph{arXiv preprint arXiv:1812.11806}}
  (\bibinfo{year}{2018}).
\newblock


\bibitem[\protect\citeauthoryear{Li, Pan, Wan, and Kot}{Li
  et~al\mbox{.}}{2019}]%
        {li2019heterogeneous}
\bibfield{author}{\bibinfo{person}{Haoliang Li}, \bibinfo{person}{Sinno~Jialin
  Pan}, \bibinfo{person}{Renjie Wan}, {and} \bibinfo{person}{Alex~C Kot}.}
  \bibinfo{year}{2019}\natexlab{}.
\newblock \showarticletitle{Heterogeneous transfer learning via deep matrix
  completion with adversarial kernel embedding}. In
  \bibinfo{booktitle}{\emph{Proceedings of the AAAI Conference on Artificial
  Intelligence}}, Vol.~\bibinfo{volume}{33}. \bibinfo{pages}{8602--8609}.
\newblock


\bibitem[\protect\citeauthoryear{Long, Cao, Wang, and Jordan}{Long
  et~al\mbox{.}}{2015}]%
        {long2015learning}
\bibfield{author}{\bibinfo{person}{Mingsheng Long}, \bibinfo{person}{Yue Cao},
  \bibinfo{person}{Jianmin Wang}, {and} \bibinfo{person}{Michael Jordan}.}
  \bibinfo{year}{2015}\natexlab{}.
\newblock \showarticletitle{Learning transferable features with deep adaptation
  networks}. In \bibinfo{booktitle}{\emph{International conference on machine
  learning}}. PMLR, \bibinfo{pages}{97--105}.
\newblock


\bibitem[\protect\citeauthoryear{Mikolov, Chen, Corrado, and Dean}{Mikolov
  et~al\mbox{.}}{2013}]%
        {mikolov2013efficient}
\bibfield{author}{\bibinfo{person}{Tomas Mikolov}, \bibinfo{person}{Kai Chen},
  \bibinfo{person}{Greg Corrado}, {and} \bibinfo{person}{Jeffrey Dean}.}
  \bibinfo{year}{2013}\natexlab{}.
\newblock \showarticletitle{Efficient estimation of word representations in
  vector space}.
\newblock \bibinfo{journal}{\emph{arXiv preprint arXiv:1301.3781}}
  (\bibinfo{year}{2013}).
\newblock


\bibitem[\protect\citeauthoryear{Shen, Dai, Chung, Lu, and Choi}{Shen
  et~al\mbox{.}}{2020a}]%
        {shen2020adversarial}
\bibfield{author}{\bibinfo{person}{Xiao Shen}, \bibinfo{person}{Quanyu Dai},
  \bibinfo{person}{Fu-lai Chung}, \bibinfo{person}{Wei Lu}, {and}
  \bibinfo{person}{Kup-Sze Choi}.} \bibinfo{year}{2020}\natexlab{a}.
\newblock \showarticletitle{Adversarial deep network embedding for
  cross-network node classification}. In \bibinfo{booktitle}{\emph{Proceedings
  of the AAAI Conference on Artificial Intelligence}},
  Vol.~\bibinfo{volume}{34}. \bibinfo{pages}{2991--2999}.
\newblock


\bibitem[\protect\citeauthoryear{Shen, Dai, Mao, Chung, and Choi}{Shen
  et~al\mbox{.}}{2020b}]%
        {shen2019network}
\bibfield{author}{\bibinfo{person}{Xiao Shen}, \bibinfo{person}{Quanyu Dai},
  \bibinfo{person}{Sitong Mao}, \bibinfo{person}{Fu-lai Chung}, {and}
  \bibinfo{person}{Kup-Sze Choi}.} \bibinfo{year}{2020}\natexlab{b}.
\newblock \showarticletitle{Network together: Node classification via
  cross-network deep network embedding}.
\newblock \bibinfo{journal}{\emph{IEEE Transactions on Neural Networks and
  Learning Systems}} \bibinfo{volume}{32}, \bibinfo{number}{5}
  (\bibinfo{year}{2020}), \bibinfo{pages}{1935--1948}.
\newblock


\bibitem[\protect\citeauthoryear{Sinha, Shen, Song, Ma, Eide, Hsu, and
  Wang}{Sinha et~al\mbox{.}}{2015}]%
        {sinha2015overview}
\bibfield{author}{\bibinfo{person}{Arnab Sinha}, \bibinfo{person}{Zhihong
  Shen}, \bibinfo{person}{Yang Song}, \bibinfo{person}{Hao Ma},
  \bibinfo{person}{Darrin Eide}, \bibinfo{person}{Bo-June Hsu}, {and}
  \bibinfo{person}{Kuansan Wang}.} \bibinfo{year}{2015}\natexlab{}.
\newblock \showarticletitle{An overview of microsoft academic service (mas) and
  applications}. In \bibinfo{booktitle}{\emph{Proceedings of the 24th
  international conference on world wide web}}. \bibinfo{pages}{243--246}.
\newblock


\bibitem[\protect\citeauthoryear{Tang, Zhang, Yao, Li, Zhang, and Su}{Tang
  et~al\mbox{.}}{2008}]%
        {tang2008arnetminer}
\bibfield{author}{\bibinfo{person}{Jie Tang}, \bibinfo{person}{Jing Zhang},
  \bibinfo{person}{Limin Yao}, \bibinfo{person}{Juanzi Li}, \bibinfo{person}{Li
  Zhang}, {and} \bibinfo{person}{Zhong Su}.} \bibinfo{year}{2008}\natexlab{}.
\newblock \showarticletitle{Arnetminer: extraction and mining of academic
  social networks}. In \bibinfo{booktitle}{\emph{Proceedings of the 14th ACM
  SIGKDD international conference on Knowledge discovery and data mining}}.
  \bibinfo{pages}{990--998}.
\newblock


\bibitem[\protect\citeauthoryear{Wang, Ji, Shi, Wang, Ye, Cui, and Yu}{Wang
  et~al\mbox{.}}{2019}]%
        {wang2019heterogeneous}
\bibfield{author}{\bibinfo{person}{Xiao Wang}, \bibinfo{person}{Houye Ji},
  \bibinfo{person}{Chuan Shi}, \bibinfo{person}{Bai Wang},
  \bibinfo{person}{Yanfang Ye}, \bibinfo{person}{Peng Cui}, {and}
  \bibinfo{person}{Philip~S Yu}.} \bibinfo{year}{2019}\natexlab{}.
\newblock \showarticletitle{Heterogeneous graph attention network}. In
  \bibinfo{booktitle}{\emph{The World Wide Web Conference}}.
  \bibinfo{pages}{2022--2032}.
\newblock


\bibitem[\protect\citeauthoryear{Wu, Pan, Zhou, Chang, and Zhu}{Wu
  et~al\mbox{.}}{2020}]%
        {wu2020unsupervised}
\bibfield{author}{\bibinfo{person}{Man Wu}, \bibinfo{person}{Shirui Pan},
  \bibinfo{person}{Chuan Zhou}, \bibinfo{person}{Xiaojun Chang}, {and}
  \bibinfo{person}{Xingquan Zhu}.} \bibinfo{year}{2020}\natexlab{}.
\newblock \showarticletitle{Unsupervised domain adaptive graph convolutional
  networks}. In \bibinfo{booktitle}{\emph{Proceedings of The Web Conference
  2020}}. \bibinfo{pages}{1457--1467}.
\newblock


\bibitem[\protect\citeauthoryear{Xiao and Guo}{Xiao and Guo}{2013}]%
        {xiao2013novel}
\bibfield{author}{\bibinfo{person}{Min Xiao} {and} \bibinfo{person}{Yuhong
  Guo}.} \bibinfo{year}{2013}\natexlab{}.
\newblock \showarticletitle{A novel two-step method for cross language
  representation learning}.
\newblock \bibinfo{journal}{\emph{Advances in Neural Information Processing
  Systems}}  \bibinfo{volume}{26} (\bibinfo{year}{2013}),
  \bibinfo{pages}{1259--1267}.
\newblock


\bibitem[\protect\citeauthoryear{Yang, Song, Jin, and Du}{Yang
  et~al\mbox{.}}{2020}]%
        {yang2020domain}
\bibfield{author}{\bibinfo{person}{Shuwen Yang}, \bibinfo{person}{Guojie Song},
  \bibinfo{person}{Yilun Jin}, {and} \bibinfo{person}{Lun Du}.}
  \bibinfo{year}{2020}\natexlab{}.
\newblock \showarticletitle{Domain adaptive classification on heterogeneous
  information networks}. In \bibinfo{booktitle}{\emph{Proceedings of the
  Twenty-Ninth International Joint Conference on Artificial Intelligence}}.
  \bibinfo{pages}{1410}.
\newblock


\bibitem[\protect\citeauthoryear{Zhang, Song, Du, Yang, and Jin}{Zhang
  et~al\mbox{.}}{2019}]%
        {zhang2019dane:}
\bibfield{author}{\bibinfo{person}{Yizhou Zhang}, \bibinfo{person}{Guojie
  Song}, \bibinfo{person}{Lun Du}, \bibinfo{person}{Shuwen Yang}, {and}
  \bibinfo{person}{Yilun Jin}.} \bibinfo{year}{2019}\natexlab{}.
\newblock \showarticletitle{Dane: Domain adaptive network embedding}. In
  \bibinfo{booktitle}{\emph{Proceedings of the Twenty-Eight International Joint
  Conference on Artificial Intelligence}}.
\newblock


\bibitem[\protect\citeauthoryear{Zhuang, Qi, Duan, Xi, Zhu, Zhu, Xiong, and
  He}{Zhuang et~al\mbox{.}}{2020}]%
        {zhuang2019a}
\bibfield{author}{\bibinfo{person}{Fuzhen Zhuang}, \bibinfo{person}{Zhiyuan
  Qi}, \bibinfo{person}{Keyu Duan}, \bibinfo{person}{Dongbo Xi},
  \bibinfo{person}{Yongchun Zhu}, \bibinfo{person}{Hengshu Zhu},
  \bibinfo{person}{Hui Xiong}, {and} \bibinfo{person}{Qing He}.}
  \bibinfo{year}{2020}\natexlab{}.
\newblock \showarticletitle{A comprehensive survey on transfer learning}.
\newblock \bibinfo{journal}{\emph{Proc. IEEE}} \bibinfo{volume}{109},
  \bibinfo{number}{1} (\bibinfo{year}{2020}), \bibinfo{pages}{43--76}.
\newblock


\end{thebibliography}

\end{document}